# Body Models in Humans and Robots

Matej Hoffmann[1] and Matthew R. Longo[2]

[1]Department of Cybernetics, Faculty of Electrical Engineering, Czech Technical University in Prague

[2]Department of Psychological Sciences, Birkbeck, University of London

*Introduction*

Neurocognitive models of higher-level somatosensory processing have emphasised the role of stored body representations in interpreting real-time sensory signals coming from the body (Longo, Azañón and Haggard, 2010; Tamè, Azañón and Longo, 2019). The need for such stored representations arises from the fact that immediate sensory signals coming from the body do not specify metric details about body size and shape. Several aspects of somatoperception, therefore, require that immediate sensory signals be combined with stored body representations. This basic problem is equally true for humanoid robots and, intriguingly, neurocognitive models developed to explain human perception are strikingly similar to those developed independently for localizing touch on humanoid robots, such as the iCub, equipped with artificial electronic skin on the majority of its body surface (Roncone *et al.*, 2014; Hoffmann, 2021). In this chapter, we will review the key features of these models, discuss their similarities and differences to each other, and to other models in the literature. Using robots as embodied computational models is an example of synthetic methodology or 'understanding by building' (e.g., Hoffmann and Pfeifer, 2018), computational embodied neuroscience (Caligiore *et al.*, 2010) or 'synthetic psychology of the self' (Prescott and Camilleri, 2019). Such models have the advantage that they need to be worked out into every detail, making any theory explicit and complete. There is also an additional way of (pre)validating such a theory other than comparing to the biological or psychological phenomenon studied by simply verifying that a particular implementation really performs the task: can the robot localize where it is being touched (see https://youtu.be/pfse424t5mQ)?

Theorizing about the body schema and its disorders is at least as old as experiments investigating tactile localization. Pick (1922) wrote: "The abundance of speculation and theoretical constructs in the doctrine of the loss of awareness of the body, i.e. of defects in the 'somatopsyche,' is inversely proportional to factual observations ..." (cited from Poeck and Orgass, 1971). More recently, the focus seems to have shifted. "These last twenty years have seen an explosion of experimental work on body representations, which should help us shape and refine our theory of body awareness." (De Vignemont, 2018) (pg. 2) Instead, theorizing has been less prominent. Figure 1 shows a recent neurocognitive model of human somatoperceptual processing (panel A) and a pipeline used with the iCub robot (panel B). In both cases, basic perceptual processes such as localisation of tactile stimuli on the body and proprioceptive localisation of the limbs in external space involve the integration of real-time afferent signals from the body with a set of stored representations of the body. In the remainder of this chapter, we will first discuss the character of the body representations proposed in Fig. 1A. Then we will review experiments tapping into the three specific perceptual abilities (shown in rectangles on the right side of each panel): somatic localisation of touch (in a skin-centred reference frame), spatial localization of touch (in external space), and spatial localization of body parts (in external space). We will then revisit and extend the original schematics (panel A), taking also the temporal dimension into account—long-term vs. short-term body representations.

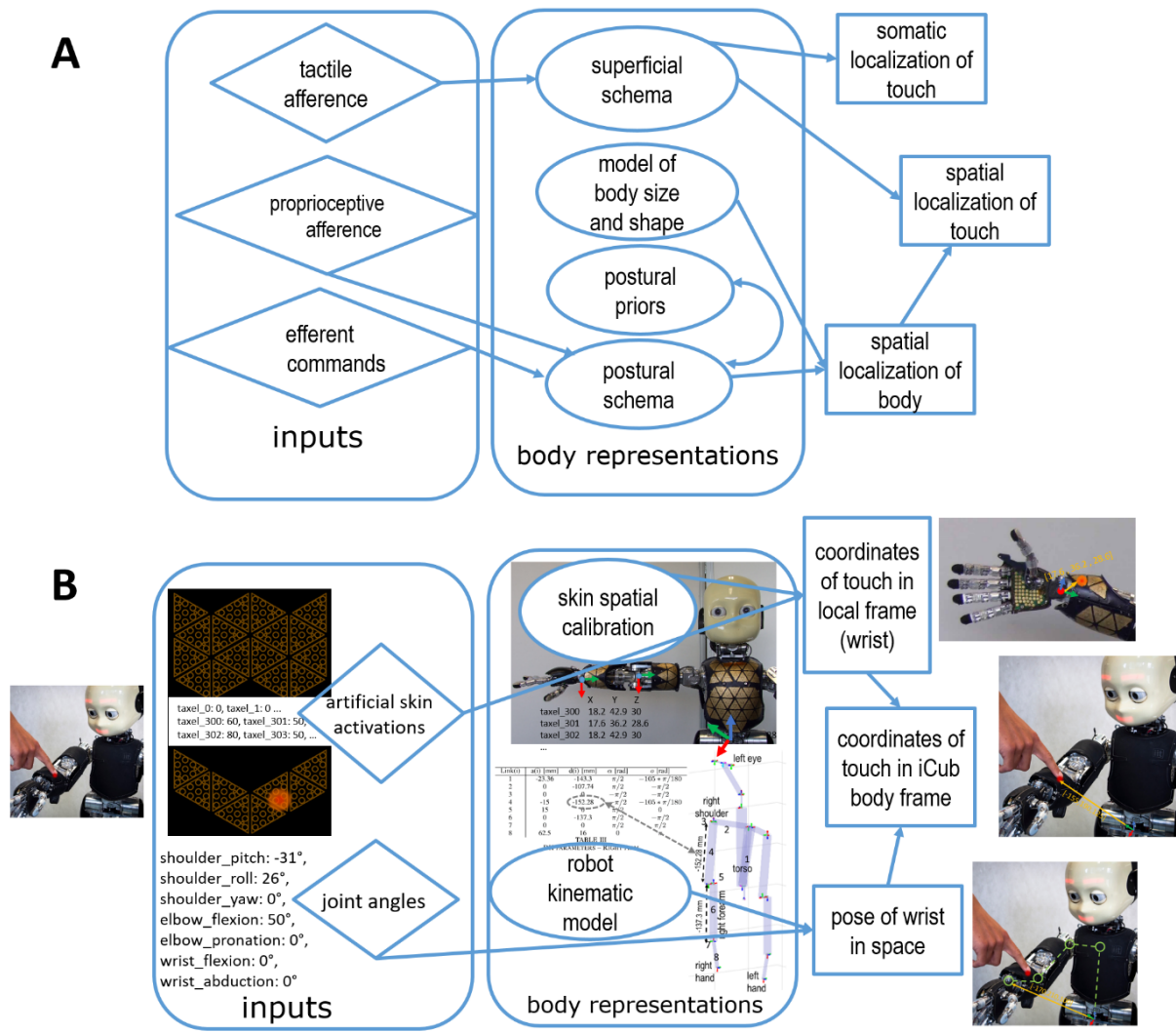

Fig. 1. (A) Conceptual model of somatosensory processing relating to spatial localization of touch (adapted from Longo, Azañón, and Haggard (2010) and Tamè, Azañón, & Longo, (2019)). (B) Spatial localization of touch on the iCub humanoid robot touched on its right forearm. *Inputs:* Touch is perceived as activations of certain taxels (i.e. tactile pixels) of the artificial skin. The joint angles of the right arm can be retrieved from encoders. *Body representations*: Positions of every taxel with respect to a local frame of reference (the wrist for skin on the forearm) are known (skin spatial calibration). The robot kinematic model is also available. *Perceptual processes*: Coordinates of touch in local frame (wrist): the positions of activated taxels can be directly retrieved from the skin spatial calibration. Pose (i.e. position and orientation) of the wrist in space can be obtained from forward kinematics—combining the robot model with current joint angle

values. Coordinates of touch in space: is obtained by adding the transformation to the activated taxel to the wrist pose.

*Postural schema and superficial schema – body models or body state representations?*

Most authors refer to Head and Holmes (1911) as the pioneers of the investigations of body schema disorders. However, Poeck and Orgass (1971) provide an excellent critical review of the investigations of the body schema up to their times, supported by tests of 50 of their neurological patients. According to them, the term *schema* as related to the orientation on the body was introduced by Bonnier (1905) who realized that beyond diffuse sensory impressions from the viscera, muscles, joints and skin ('coenesthesia'), there must be assumed some spatial quality in the awareness of the body. The motivation for postulating these representations in the first place comes from the fact that afferent sensory signals do not provide direct information about body size and shape and hence some body models are necessary: real-time sensory signals have to be combined with stored information about the spatial properties of the body.

The notions of *postural schema* and *superficial schema* are attributed to Head and Holmes (1911). These particular labels never appear in their original work, but they discuss them. Regarding the postural schema, they write: "*By means of perpetual alterations in position we are always building up a postural model of ourselves which constantly changes. Every new posture or movement is recorded on this plastic schema, and the activity of the cortex brings every fresh group of sensations evoked by altered posture into relation with it.*" The superficial schema, in turn, is defined as follows: "*In the same way, recognition of the locality of the stimulated spot demands the reference to another "schema." For a patient may be able to name correctly, and indicate on a diagram or on another person's hand, the exact position of the spot touched or pricked, and yet be ignorant of the position in space of the limb upon which it lies.*"

More recently, in neurocognitive models of somatoperception (Longo et al., 2010; Tamè et al., 2019), the superficial schema is thought of as a lookup table, which maps locations of neural activation within a somatotopic map onto locations defined relative to a structural representation of the skin surface. The postural schema consists of a constantly-updated collection of joint angles specifying the current posture of the body, which can be combined with the stored body model to determine absolute location of body parts in space. The postural schema receives inputs from afferent proprioceptive signals, from efferent copies of motor commands, and also from postural priors (Romano et al., 2017; 2019) which serve as Bayesian priors for perceived posture.

Let us now introduce the temporal dimension based on the dynamics of body representations distinguishing long-term, or offline, and short-term, or online, body representations (O'Shaughnessy, 1995; Carruthers, 2008). It is not clear whether the schemas refer to the long-term body models (this seems to be the case for the superficial schema in the form of a lookup table) or body state representations (postural schema as a constantly updated collection of joint angles). We will use this additional dimension to revise the model.

*Somatosensory localization – experimental evidence*
In this section, we review selected experiments investigating the three key somatoperceptual processes of Fig. 1A: somatic localization of touch, spatial localization of touch, and spatial localization of the body. Different tasks have been devised to study them. We review the localization patterns on different variants of these tasks, paying specific attention to interactions or dissociations of the processes, which importantly informs our modelling endeavour.

*Somatic versus spatial localization of touch.* The first somatoperceptual ability we will discuss is perceiving the location of a touch on the body. For example, if someone touches one of your fingers, it is straightforward to judge which finger it was that was touched. However, detecting touch does not automatically mean localizing it. Head and Holmes (1911) reported a number of brain-damaged patients who could detect touch, but were unable to localise it on the body. For example, one patient described their experience (p. 139): "I feel you touch me, but I can't tell where it is; the touch oozes all through my hand." That these patients were able to detect touch indicated that the initial sensory processing of tactile information was intact, and that what was disrupted was a subsequent localisation stage of processing. Other patients show systematic deformations of the mapping of actual touch location on perceived location (Rapp et al., 2002). Finally, given plasticity of primary somatotopic maps as a result of experience (e.g., Merzenich et al., 1984) and expertise (e.g., Elbert et al., 1995; Pascual-Leone & Torres, 1993), location on the skin cannot be equated with location within a somatotopic map.

    As suggested by Fig. 1A, it seems that one should further distinguish localizing touch in a skin-centred reference frame as opposed to localizing it in external space. Targeted experiments can help uncover how these two perceptual processes may be supported by distinct or common internal models of the body. A simple test of tactile localisation involves stimulating the subjects on their hand or forearm with the stimulus site out of sight and after removal of the stimulus, asking them to reach or point to the stimulated location (Paillard, Michel and Stelmach, 1983; Rossetti, Rode and Boisson, 1995; Paillard, 1999; Rapp, Hendel and Medina, 2002), which may be also hidden under an occluding board (Longo, Mancini and Haggard, 2015; Medina and Duckett, 2017) or touch screen (Mueller and Fiehler, 2014) on which the location of the stimulus is also indicated—for example, by pointing using a baton (Longo and Morcom, 2016; Longo, 2017). Importantly, performing the task requires localizing the touch in external space.

    An alternative family of localization tests involves localizing the tactile stimulus on a grid drawn on the subject's limb (Weinstein, 1968), pointing on a drawing (Rossetti, Rode and Boisson, 1995; Paillard, 1999), a downsized drawing on a computer screen (Miller *et al.*, 2020), indicating touch locations on a silhouette shape of the body (Head, 1920; Mancini *et al.*, 2011; Medina, Tamè and Longo, 2018), or on the experimenter's hand (Head and Holmes, 1911). In these cases, reaching towards the touched location on the subject's body is not involved and hence, it may be possible to localize touch without localizing the stimulated body part in space.

    There are localization patterns that seem independent from the particular way of localizing touch. Most importantly, localization is more accurate near landmarks (typically joints on the body), and less accurate around the centre of body parts – e.g., the (ventral/volar) forearm (Cholewiak and Collins, 2003; Miller *et al.*, 2020). In the study of Miller and colleagues (2020), this profile is apparent in several variants of touch localization (external task / drawing task, passive touch / active touch). The results from the tactile task in Longo (2017) seem to confirm better accuracy in the proximo-distal axis, but bigger errors in the medio-lateral axis around the wrist. On the other hand, there are localization patterns specific to how the task is performed. Mancini et al. (2011), using non-manual localization, found systematic distal and radial shift (toward the thumb) in localization on the hairy skin of the hand dorsum. Such

biases were not found for the glabrous skin of the palm. These biases were not found when pointing to the stimulated point on an occluding board in Longo et al. (2015).

*Localising body landmarks in external space.* The second somatoperceptual ability we will discuss is the ability to proprioceptively perceive the location of body parts in external 3-dimensional space (spatial localisation of body in Fig. 1A). Head and Holmes (1911) reported patients who could perceive touch and localise it on the skin, but could not tell where their arm was in external space if they couldn't see it. Indeed, Head and Holmes went so far as to declare that "Inability to recognize the position of the affected part in space is the most frequent sensory defect produced by lesions of the cerebral cortex" (pg. 157). A similar setup that was used for tactile localization in Longo & Morcom (2016) and Longo (2017b) can be also used to probe 'proprioceptive maps'. Participants are asked to mark on an occluding board the locations of specific body landmarks—fingertips and knuckles of the hand, for example. Longo and Haggard (2010) and Longo (2017b) (verbal task) found the following patterns of distortion of the perceptual maps: (i) overall underestimation of finger length; (ii) a radial-ulnar gradient with underestimation increasing from the thumb-side to the little finger-side of the hand, and (iii) overestimation of hand width. Similar patterns were also replicated by Mattioni and Longo (2014) when tactile cueing was used in addition to verbal cuing.

*Interactions between localising touch and body.* Finally, to further tap into the mechanisms of somatoperceptual processing, researchers have also combined some of the tasks reported above. Such experiments are critical to understand the interplay of the body representations enabling the individual somatoperceptual processes. Longo and Morcom (2016) combined the tactile localization task and tactile distance perception. Tactile stimuli localized individually were used to compute perceived distances and compared with explicit distance judgments made by the participants. During explicit tactile distance perception, there was a clear bias to perceive distances across the width of the hand as bigger than those along the length of the hand. When inferred from individual localizations, significant overestimation was found for all distances, with the magnitude of overestimation significantly larger in the across than in the along orientation in all cases. However, statistical correlations between the tasks were not found, which may suggest that to accomplish these tasks, at least partially separate body representations may be recruited.

    The tactile localization task and the body landmark localization were combined in the occluding board setup in Longo, Mancini & Haggard (2015) with localizations on the hand dorsum. In addition, localization of touch on a hand silhouette like in Mancini et al. (2011) was part of the same study. The distortions characteristic of 'position sense' mentioned above were also seen in tactile localization when participants had to point to the stimulated point on the hand dorsum (on the occluding board). However, the biases of tactile localization on the hand silhouette found in Mancini et al. (2011) and replicated in Longo et al. (2015)—distal and radial bias—were not found for localization when pointing. This is suggestive of a possible dissociation of the processes—somatic localization of touch in a skin-based frame may not flow into spatial localization of touch (see Fig. 1A).

*Predictions and implications for a process model*
A large amount of empirical evidence regarding somatosensory localization and distance perception has been accumulated so far. The evidence is seldom integrated into a computational or at least process model that would suggest possible mechanisms, integrating the evidence and generating new testable predictions. The conceptual diagram in Fig. 1A constitutes such an attempt. Its evolution (Longo, Azañón and Haggard, 2010; Longo, Mancini and Haggard, 2015; Tamè, Azañón and Longo, 2019) reflects the integration of new evidence. Three different perspectives on tactile remapping are also graphically depicted in Heed et al. (2015). Still, such conceptualizations remain very high-level.

Several points can be drawn from the empirical evidence. First, the accuracy of somatosensory localization is limited. Not only are localisation responses noisy and variable, but they also show a range of systematic biases and distortions.

Second, specific properties of the somatosensory system—receptive field properties and distortions of the sensory homunculus—appear to be partially reflected in somatosensory localization processes. Yet they are importantly attenuated or compensated for. For example, Taylor-Clarke et al. (2004) estimated that Weber's illusion is just 10% of what would be predicted from differences in tactile acuity alone.

Third, it remains unclear to what extent there is a single somatoperceptual processing 'pipeline'—like the one in Fig. 1A—that is recruited for all the tasks related to somatosensory localization. For example, there are different ways of performing tactile localization like pointing to the stimulated point or marking it on a model or silhouette of the body part. Some patterns of (mis)localization seem to hold across tasks (see e.g., the U-shaped profile reflecting better accuracy near landmarks reported in several localization tasks in Miller et al. (2020)) while others seem specific to the task (see Longo et al. 2015).

Fourth, the empirical evidence, though abundant, focuses largely on hands and forearms. Importantly, the hand or forearm almost always lies flat on a table and the joints of the wrist or elbow are not moved. This represents a major limitation of the literature, as incorporating the signals from the joints posture seems critical for tactile remapping and its understanding. Recent studies have started to address this gap, applying many of the tasks described above to other parts of the body, including the face (e.g., Mora *et al.*, 2018; Longo and Holmes, 2020; Longo *et al.*, 2020), legs (e.g., Stone, Keizer and Dijkerman, 2018; Tosi and Romano, 2020), feet (e.g., Cicmil, Meyer and Stein, 2016; Manser-Smith, Tamè and Longo, 2018, 2021), and torso (e.g., Longo, Lulciuc and Sotakova, 2019; Plaisier, Sap and Kappers, 2020; Nicula and Longo, 2021).

*Conceptual model of somatosensory processing versus robot architecture*
Let us now have a close look at the differences between Fig. 1 A and B. Note that Fig. 1A is a conceptual model while the robot pipeline (Fig. 1B) is an actual implementation and we know that it works and we also know exactly how it works.

First, the 'wiring' is different. In the robot architecture, inputs (diamond shapes) + body representations (circles) give rise to the perceptual processes (rectangles). For example, there is a skin calibration table (body representation), in which 3D coordinates of every tactile receptor in a local frame of reference are stored. With skin activation coming (inputs), the coordinates of the activated taxel (i.e., tactile pixel) can be simply looked up, giving rise to *coordinates of touch in the local frame*—with respect to the wrist in this case (top right in Fig. 1B). To localize the wrist in space (*pose of wrist in space*), the robot kinematic model is deployed in the forward kinematics function which takes the joint angles as input. The robot has its centre of coordinates in its waist. To localize the wrist in space with respect to this body-centred frame, a sequence of matrix multiplications along the kinematic chain depicted in green in Fig. 1B, bottom right, with the current joint angles as parameters, gives rise to a 3D vector from the body frame to the wrist (see Hoffmann, 2021 for more details).

In Fig. 1A, the inputs are not combined with but flow 'through' the representations—the *superficial schema* and the *postural schema*. To what extent the superficial schema may be a lookup table with 'coordinates' of receptors and how that may operate should be studied in detail. Regarding the postural schema, note that in Fig. 1B, there is no such module. The *robot kinematic model* seems to correspond to the *model of body size and shape* in Fig. 1A. What role would the postural schema play then? The *model of body size and shape / robot kinematic model* constitutes a long-term or offline body representation, the content of which is expected to be relatively stable in adulthood—what the body is usually like. Overall, the nature of the *superficial* and *postural* schema is not clear. Possibly, they may correspond to short-term body representations, representing body states, carrying the outcome of the actual perceptual processes encoded in them.

Second, to bring about spatial localization of touch, i.e. tactile remapping, in the robot architecture (Fig. 1B), the coordinates of the wrist with respect to the body frame are obtained (*pose of wrist in space*) and the 'last bit' from the wrist to the activated skin (*coordinates of touch in local frame (wrist)*) is added. Note that to go from the body frame to the wrist, a full 6D transformation is needed—a 3D translation and a 3D rotation of the reference frame. The orientation is needed such that the last translation to the stimulated taxel is added in the right direction to the wrist frame. This architecture is a parsimonious way of achieving tactile remapping. However, in the conceptual diagram of Fig. 1A, the superficial schema independently feeds the *somatic localization of touch* and the *spatial localization of touch*, based on the evidence discussed above (Longo et al. 2015 in particular). This wiring seems redundant from an engineering standpoint, but redundancy is common in biological systems.

Third, the conceptual schematics in Fig. 1A features predictive mechanisms—efferent copies of motor commands and postural priors—that feed the postural schema. There is no conceptual or technical reason why these could not be used in robots, but they are not necessary. Proprioceptive position sense in humans is notoriously noisy (Tillery, Flanders and Soechting, 1991) and subject to delays in the corresponding ascending pathways from the receptors like muscle spindles to the brain. Estimation of the body state can thus be improved by including efference copies of motor commands as well as priors—typical body postures. These may operate locally, like for frequent joint configurations, or more globally like for habitual postures of the whole body. Such priors serve two functions: (i) they improve the estimation (in the Bayesian sense, for example); (ii) they may substitute for actual perception ahead of time (see also Azañón and Soto-Faraco, 2008). In robots, on the other hand, proprioceptive sense in the form of joint encoders is typically both accurate and fast. However, if no joint angle values are fed into the forward kinematics function, it will automatically assume the default position of every joint (imagine every joint angle equal to 0 degrees), giving rise to a canonical robot body posture, potentially analogous to the idea of a standard body posture in humans (cf. Romano *et al.*, 2019).

*Conceptual model of somatosensory processing revisited*
Here we attempt to integrate the insights from the previous sections into a new revision of the conceptual model (Fig. 2). The main difference compared to Fig. 1A is that we newly distinguish between long-term body representations (body models) and short-term representations of body states. The short-term body state representations can be best viewed as neural populations that through their activity encode certain extraneural body states. We also attempt to assign putative brain areas to these blocks. The long-term body representations are necessary for the respective transformations between blocks. However, these body models may not explicitly exist anywhere but may be implicitly realized in distributed patterns of synaptic connections, for example. This is a typical property expected from neural representations in the tradition of parallel distributed processing, also known as connectionism (Rumelhart and McClelland, 1986); see also Ramsey (2007). Hence, the inputs from the body models are marked with dashed arrows in Fig. 2. The flow and mixing of tactile and proprioceptive information is inspired by what is known from neurophysiology (see, e.g., Dijkerman and de Haan, 2007, for a review). Efference copies are speculatively marked as flowing into the postural schema.

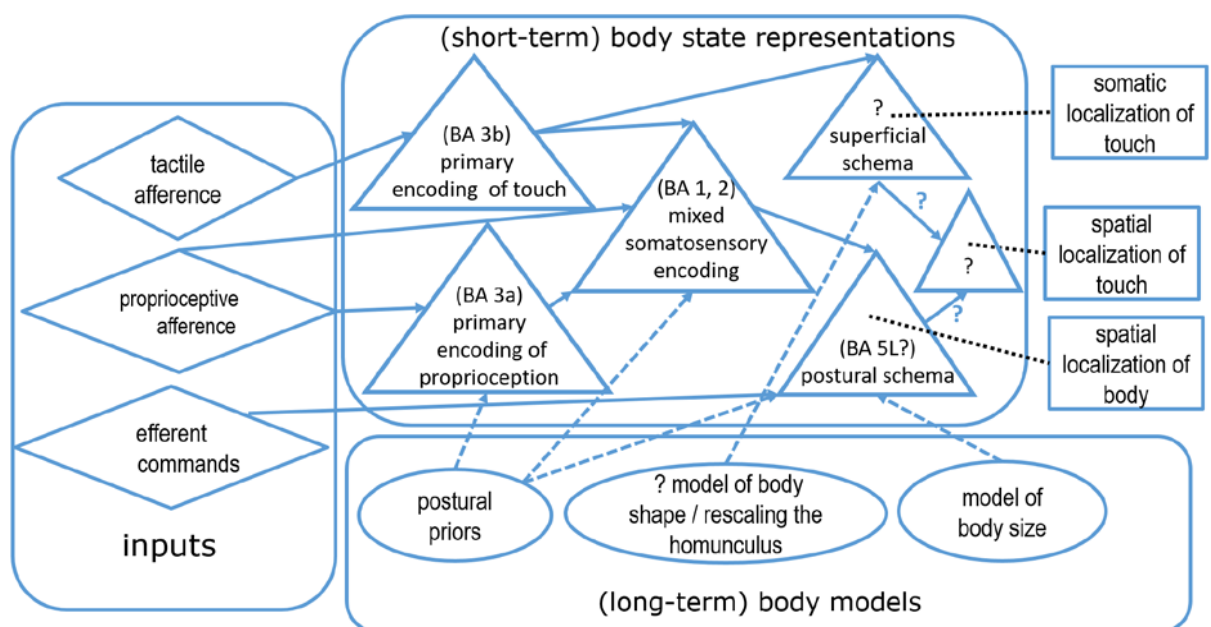

Fig. 2. Revisited Conceptual model of somatosensory processing relating to spatial localization of touch. The contribution of the long-term body models to the pipeline is marked with dashed arrows. In reality, their effect may be embedded in the strengths of the synaptic connections (full arrows). Some short-term body state representations (triangles) toward the right of the schematics may encode the outcome of specific perceptual processes (rectangles). Putative brain areas are assigned to some representations (BA – Brodmann area). Question marks are used whenever the existence or role of a certain block or connection is not clear.

The postural schema is a putative site of somatosensory representations after remapping—possibly 'storing' the location of body parts in a body-centred frame of reference, for example. Hence, this population, postural schema, may be the site where one could 'read out' the spatial localization of the body. There is a clear correspondence with the robot architecture in Fig. 1B—the *model of body size* corresponding to the *robot kinematic model. Postural priors* can in principle be present at any stage of processing and may be expressed in more local and anatomical coordinates (Brodmann areas 3a, 1, 2) or perhaps in body configurations in space (area 5L in PPC).
  The placement and nature of the superficial schema remains less clear. While the postural schema would be the site of somatosensory remapping, the superficial schema may reflect a rescaled version of the tactile homunculus (area 3b) where the distortions have been attenuated. The evidence from Mancini et al. (2011) and Longo et al. (2015) reviewed above reports specific biases related to localization of touch in a skin-based frame of reference, without the need for tactile spatial remapping. If such a population existed, it would need to receive the information from a *model of body shape*, which one can picture as the rescaling realized through synaptic connections, for example. This would correspond to the *skin spatial calibration* in Fig.1B—coordinates of tactile receptors in a local frame of reference. However, it seems still unclear whether a separate pipeline dealing with only tactile information exists.

To achieve spatial localization of touch, the information from the postural and superficial schema may be combined, just like in the robot architecture (Fig. 1B). However, it is not clear whether there is a separate neural population encoding this and whether it is fed by the superficial schema. In our schematics, the postural schema is already fed with tactile information from Brodmann areas 1 and 2 and hence, this block may already be in a position to achieve not only spatial localization of body but also of touch.

Here (Fig. 2) we propose to think of the postural schema as a short-term representation of the current body posture in spatial (e.g., body-centred) coordinates. Inspired by the robot architecture (Fig. 1B), where the position and orientation of all joints with respect to the body frame can be obtained from forward kinematics combining the robot kinematic model and current joint angle values, the postural schema would similarly consist of the coordinates of body landmarks (joints) in a body-centred frame, obtained by combining the *model of body size* and the joint angle values. With respect to the previous versions (Fig. 1A), we kept only the *body size* from the *model of body size and shape*, since for this operation, the full shape of the surface of the body is not necessary. The postural schema contains merely a 'stick figure' in the current body posture. Differently to the robot world, the joint angle values are harder to obtain. Hence, proprioceptive afference is complemented with efference copies (inputs) and postural priors—canonical positions of individual joints or the whole body (Romano, Marini and Maravita, 2017; Romano *et al.*, 2019) regarding standard body posture. We propose area 5L in the posterior parietal cortex as a potential site of this neural population. Azañón et al. (2010) wrote: "Postural schema and canonical posture would therefore play roles similar to 'current evidence' and 'prior probability' terms, respectively". In our account here, we conceive of the postural schema rather as a population holding the 'posterior' about the body configuration, integrating the postural priors and current sensory information. Unlike the afferent information, the postural priors may be available almost instantly and hence initial body estimation may be biased toward the canonical posture (see Azañón and Soto-Faraco, 2008).

The superficial schema seems more challenging. In the robot architecture (Fig. 1B), localizing on the skin in a local frame of reference is achieved by simply looking up the coordinates of an activated skin receptor in a *skin spatial calibration* table. The coordinates of every receptor on a body part form a 3D point cloud which implicitly defines the shape of the body part. Hence, we kept the *body shape* as a separate long-term body representation from the *body size*. It remains to be defined whether the superficial schema would correspond to such a 'lookup table' (a long-term body representation) or rather the representation of current state—current 'coordinates' of touch in a local frame of reference. This choice seems arbitrary; in Fig. 2, for consistency with the postural schema, the superficial schema is conceived as a short-term body representation. However, the evidence about the existence of such a representation rescaling the distorted representation of the skin surface (the homunculus, area 3b) without remapping it into external space seems inconclusive. Similarly, the site in the brain is unclear (and hence no brain area assigned in Fig. 2 to the superficial schema).

*Conclusion, Discussion and Future work*
In this chapter, we attempted to revise the earlier conceptual models of somatosensory processing relating to spatial localization of touch (Longo, Azañón, and Haggard (2010) and Tamè, Azañón, & Longo, (2019)). The goal of this exercise was to, first, add additional details and bring the highly abstract schematics close to a process model that could potentially be later transformed into a computational model. Second, taking into account existing evidence about the neural realization of these processes, we assigned possible brain sites to the respective blocks. The implementation of localizing stimuli on the body of a humanoid robot (Roncone et al. 2014; Hoffmann 2021) served as inspiration and an example of in some sense ideal, engineered, way of realizing the necessary operations.

The main contributions of the new conceptualization are the following. First, we added the temporal dimension and distinguished between short-term, online, and long-term, offline, body representations (body state representation vs. body model). The former can be viewed as neural activations in a certain brain area; the latter may be implicitly encoded in synaptic weights—in line with the nature of representations in connectionist models in general (Rumelhart and McClelland, 1986). Second, we separated the model of body size from the model of body shape. To localize body landmarks (joints) in space, a stick figure seems sufficient. Instead, to localize touch, the layout of the receptors on a body part—body shape—seems necessary. Third, the exact nature and perhaps even the existence of the superficial schema as a representation dealing solely with tactile inputs seems unclear. It remains to be tested whether there is pure rescaling of the distorted skin representations without remapping them into space with the help of proprioception. Fourth, it is also not clear to what extent there would be a universal somatosensory processing pipeline like the one shown in Fig. 2 that would feed all the different tasks requiring the localization of body and touch. Finally, the robot architecture shows that a single landmark (e.g. wrist) suffices to localize touch on, say, the forearm. The neural surveyor model (Miller et al. 2020) uses triangulation between two landmarks (e.g., wrist and elbow) surrounding the corresponding skin part (forearm). More experimental evidence in favour of one or the other model is needed.

In this chapter, the role of visual inputs and visual body representations was not considered. While localization of touch is theoretically possible completely without visual inputs, in the normal brain, it is likely to be influenced by vision and representations of the body that have a predominantly visual origin. This is demonstrated for example by visual illusions altering tactile distance perception (Taylor-Clarke, Jacobsen and Haggard, 2004) or by the effects of gaze on touch localization (Harrar and Harris, 2009; Mueller and Fiehler, 2014; Medina, Tamè and Longo, 2018). It is also possible that some of the tactile localization paradigms that attempt to isolate somatic localization of touch by referring to a silhouette of a hand on a screen involve some representations with a visual origin (body image). Integrating these links to the process models presented here remains our future work.

Regarding future experimental work that would further inform the modelling endeavour, we suggest the following types of experiments. First, the incorporation of proprioception is key for remapping. However, due to practical or methodological issues, to our knowledge, there is no work where localization would be performed in several postures and the joint angles (of e.g., wrist and elbow) systematically manipulated. Second, almost all localization experiments are performed such that a stimulus is presented, then removed, and then the subject performs localization. It seems that a sustained stimulus would be more natural and will circumvent the 'tactile memory' factor. Similarly, active tactile search (Fuchs et al. 2020) also constitutes a more naturalistic setting. Third, more attention should be paid to the way of localization—whether it involves manual action, for example, or localization on a silhouette (like Longo et al. 2015)—to clarify to what extent there is a shared somatoperceptual processing pipeline.

*Acknowledgements*
We would like to thank Adrian Alsmith and Jason Khoury for comments on this manuscript. M.H. was supported by the Czech Science Foundation (GA CR), project EXPRO (no. 20-24186X).

References

Azañón, E. *et al.* (2010) 'The posterior parietal cortex remaps touch into external space', *Current Biology*, 20(14), pp. 1304–1309. doi: 10.1016/j.cub.2010.05.063.

Azañón, E. and Soto-Faraco, S. (2008) 'Changing reference frames during the encoding of tactile events.', *Current Biology*, 18, pp. 1044–1049. doi: 10.1016/j.cub.2008.06.045.

Bonnier, P. (1905) 'L'Aschématie', *Revue Neurologique*, 13, pp. 604–609.

Caligiore, D. *et al.* (2010) 'TRoPICALS: A computational embodied neuroscience model of compatibility effects', *Psychological Review*, 117, pp. 1188–1228. doi: 10.1037/a0020887.

Carruthers, G. (2008) 'Types of body representation and the sense of embodiment', *Consciousness and cognition*, 17, pp. 1302–1316. doi: 10.1016/j.concog.2008.02.001.

Cholewiak, R. W. and Collins, A. A. (2003) 'Vibrotactile localization on the arm: Effects of place, space, and age.', *Perception & Psychophysics*, 65, pp. 1058–1077.

Cicmil, N., Meyer, A. P. and Stein, J. F. (2016) 'Tactile toe agnosia and percept of a "missing toe" in healthy humans', *Perception*, 45, pp. 265–280. doi: 10.1177/0301006615607122.

Dijkerman, H. C. and de Haan, E. H. F. (2007) 'Somatosensory processes subserving perception and action.', *Behavioral and Brain Sciences*, 30, pp. 189–201. doi: 10.1017/S0140525X07001392.

Harrar, V. and Harris, L. R. (2009) 'Eye position affects the perceived location of touch', *Experimental Brain Research*, 198, pp. 403–410. doi: 10.1007/s00221-009-1884-4.

Head, H. (1920) *Studies in neurology*. London: Oxford University Press.

Head, H. and Holmes, G. (1911) 'Sensory disturbances from cerebral lesions', *Brain*, 34, pp. 102–254.

Heed, T. *et al.* (2015) 'Tactile remapping: From coordinate transformation to integration in sensorimotor processing', *Trends in Cognitive Sciences*, 19, pp. 251–258.

Hoffmann, M. (2021) 'Body models in humans, animals, and robots: Mechanisms and plasticity', in Ataria, Y., Tanaka, S., and Gallagher, Shaun (eds) *Body Schema and Body Image: New Directions*. Oxford University Press, pp. 152–180.

Hoffmann, M. and Pfeifer, R. (2018) 'Robots as powerful allies for the study of embodied cognition from the bottom up', in Newen, A., de Bruin, L., and Gallagher, S. (eds) *The Oxford Handbook 4e Cognition*. Oxford University Press, pp. 841–862.

Longo, M. R. (2017) 'Expansion of perceptual body maps near – but not across – the wrist', *Frontiers in Human Neuroscience*, 11(March), pp. 1–9. doi: 10.3389/fnhum.2017.00111.

Longo, M. R. *et al.* (2020) 'Anisotropies of tactile distance perception on the face', *Attention, Perception, & Psychophysics*, 82, pp. 3636–3647. doi: 10.3758/s13414-020-02079-y.

Longo, M. R., Azañón, E. and Haggard, P. (2010) 'More than skin deep: Body representation beyond primary somatosensory cortex.', *Neuropsychologia*, 48, pp. 655–668. doi: 10.1016/j.neuropsychologia.2009.08.022.

Longo, M. R. and Haggard, P. (2010) 'An implicit body representation underlying human position sense.', *Proceedings of the National Academy of Sciences*, 107, pp. 11727–11732. doi: 10.1073/pnas.1003483107.

Longo, M. R. and Holmes, M. (2020) 'Distorted perceptual face maps', *Acta Psychologica*. Elsevier, 208, p. 103128. doi: 10.1016/j.actpsy.2020.103128.

Longo, M. R., Lulciuc, A. and Sotakova, L. (2019) 'No evidence of tactile distance anisotropy on the belly', *Royal Society Open Science*, 6, p. 180866. doi: 10.1098/rsos.180866.

Longo, M. R., Mancini, F. and Haggard, P. (2015) 'Implicit body representations and tactile spatial remapping', *Acta Psychologica*, 160, pp. 77–87. doi: 10.1016/j.actpsy.2015.07.002.

Longo, M. R. and Morcom, R. (2016) 'No correlation between distorted body representations underlying tactile distance perception and position sense', *Frontiers in Human Neuroscience*, 10, p. 593. doi: 10.3389/fnhum.2016.00593.

Mancini, F. *et al.* (2011) 'A supramodal representation of the body surface', *Neuropsychologia*. Elsevier Ltd, 49, pp. 1194–1201. doi: 10.1016/j.neuropsychologia.2010.12.040.

Manser-Smith, K., Tamè, L. and Longo, M. R. (2018) 'Tactile confusions of the fingers and toes', *Journal of Experimental Psychology: Human Perception and Performance*, 44, pp. 1727–1738. doi: 10.1037/xhp0000566.

Manser-Smith, K., Tamè, L. and Longo, M. R. (2021) 'Tactile distance anisotropy on the feet', *Attention, Perception, & Psychophysics*.

Mattioni, S. and Longo, M. R. (2014) 'The effects of verbal cueing on implicit hand maps', *Acta Psychologica*, 153, pp. 60–65. doi: 10.1016/j.actpsy.2014.09.009.

Medina, J. and Duckett, C. (2017) 'Domain-general biases in spatial localization: Evidence against a distorted body model hypothesis', *Journal of Experimental Psychology: Human Perception and Performance*, 43, pp. 1430–1443. doi: 10.1037/xhp0000397.

Medina, S., Tamè, L. and Longo, M. R. (2018) 'Tactile localization biases are modulated by gaze direction', *Experimental Brain Research*. Springer Berlin Heidelberg, 236(1), pp. 31–42. doi: 10.1007/s00221-017-5105-2.

Miller, L. E. *et al.* (2020) 'A neural surveyor in somatosensory cortex', *bioRxiv*. doi: 10.1101/2020.06.26.173419.

Mora, L. *et al.* (2018) 'My true face: Unmasking one's own face representation', *Acta Psychologica*, 191, pp. 63–68. doi: 10.1016/j.actpsy.2018.08.014.

Mueller, S. and Fiehler, K. (2014) 'Effector movement triggers gaze-dependent spatial coding of tactile and proprioceptive-tactile reach targets', *Neuropsychologia*. Elsevier, 62, pp. 184–193. doi: 10.1016/j.neuropsychologia.2014.07.025.

Nicula, A. and Longo, M. R. (2021) 'Tactile distance perception on the back', *Perception*.

O'Shaughnessy, B. (1995) 'Proprioception and the body image', in Bermúdez, J. L., Marcel, A. J., and Eilan, N. (eds) *The body and the self*. MIT Press, pp. 175–203.

Paillard, J. (1999) 'Body schema and body image - A double dissociation in deafferented patients', in Gantchev, G. N., Mori, S., and Massion, J. (eds) *Motor control, today and tomorrow*. Academic, pp. 197 – 214.

Paillard, J., Michel, F. and Stelmach, G. (1983) 'Localization without content: A tactile analogue of "blind sight"', *Archives of Neurology*, 40, pp. 548–551. doi: 10.1001/archneur.1983.04050080048008.

Pick, A. (1922) 'Störung der Orientierung am eigenen Körper. Beitrag zur Lehre vom Bewusstsein des eigenen Körpers', *Psychologische Forschung*, 1, pp. 303–318. doi: 10.1007/BF00410392.

Plaisier, M. A., Sap, L. I. N. and Kappers, A. M. L. (2020) 'Perception of vibrotactile distance on the back', *Scientific Reports*. Nature Publishing Group UK, 10, p. 17876. doi: 10.1038/s41598-020-74835-x.

Poeck, K. and Orgass, B. (1971) 'The concept of the body schema: A critical review and some experimental results', *Cortex*, 7, pp. 254–277. doi: 10.1016/S0010-9452(71)80005-9.

Prescott, T. J. and Camilleri, D. (2019) 'The synthetic psychology of the self', in Ferreira,

M. I. A., Sequeira, J. S., and Ventura, R. (eds) *Cognitive Architectures. Intelligent Systems, Control and Automation: Science and Engineering*. Springer, pp. 85–104.

Ramsey, W. M. (2007) *Representation reconsidered*. Cambridge University Press.

Rapp, B., Hendel, S. K. and Medina, J. (2002) 'Remodeling of somotasensory hand representations following cerebral lesions in humans.', *Neuroreport*, 13(2), pp. 207–11. Available at: http://www.ncbi.nlm.nih.gov/pubmed/11893911.

Romano, D. *et al.* (2019) 'The standard posture of the hand', *Journal of Experimental Psychology: Human Perception and Performance*, 45, pp. 1164–1173. doi: 10.1037/xhp0000662.

Romano, D., Marini, F. and Maravita, A. (2017) 'Standard body-space relationships: Fingers hold spatial information', *Cognition*, 165, pp. 105–112. doi: 10.1016/j.cognition.2017.05.014.

Roncone, A. *et al.* (2014) 'Automatic kinematic chain calibration using artificial skin: Self-touch in the iCub humanoid robot', in *IEEE International Conference on Robotics and Automation (ICRA)*, pp. 2305–2312.

Rossetti, Y., Rode, G. and Boisson, D. (1995) 'Implicit processing of somaesthetic information: A dissociation between where and how?', *NeuroReport*, 6, pp. 506–510.

Rumelhart, D. E. and McClelland, J. L. (1986) *Parallel distributed processing: Explorations in the microstructure of cognition*. MIT Press.

Stone, K. D., Keizer, A. and Dijkerman, H. C. (2018) 'The influence of vision, touch, and proprioception on body representation of the lower limbs', *Acta Psychologica*. Elsevier, 185, pp. 22–32. doi: 10.1016/j.actpsy.2018.01.007.

Tamè, L., Azañón, E. and Longo, M. R. (2019) 'A conceptual model of tactile processing across body features of size, shape, side, and spatial location', *Frontiers in Psychology*, 10, p. 291. doi: 10.3389/fpsyg.2019.00291.

Taylor-Clarke, M., Jacobsen, P. and Haggard, P. (2004) 'Keeping the world a constant size: Object constancy in human touch', *Nature Neuroscience*, 7, pp. 219–220. doi: 10.1038/nn1199.

Tillery, S. I. H., Flanders, M. and Soechting, J. F. (1991) 'A coordinate system for the synthesis of visual and kinesthetic information', *Journal of Neuroscience*, 11, pp. 770–778.

Tosi, G. and Romano, D. (2020) 'The longer the reference, the shorter the legs: How response modality affects body perception', *Attention, Perception, & Psychophysics*. Attention, Perception, & Psychophysics, 82, pp. 3737–3749. doi: 10.3758/s13414-020-02074-3 The.

Trojan, J. *et al.* (2010) 'Spatiotemporal integration in somatosensory perception: effects of sensory saltation on pointing at perceived positions on the body surface', *Frontiers in Psychology*, 1, p. 206. doi: 10.3389/fpsyg.2010.00206.

Weinstein, S. (1968) 'Intensive and extensive aspects of tactile sensitivity as a function of body part, sex, and laterality', in Kenshalo, D. R. (ed.) *The skin senses*. Springfield, IL: Thomas, pp. 195–222.